\documentclass[]{ecai}
\usepackage{graphicx}
\usepackage{latexsym}

\usepackage[utf8]{inputenc}
\usepackage{alphabeta}
\usepackage{balance}
\usepackage{algorithm}
\usepackage{algpseudocode}
\usepackage{enumitem}
\usepackage{textcomp}
\usepackage{amsmath}
\usepackage[font=scriptsize]{subcaption}
\usepackage[textwidth=1.2cm, textsize=tiny, disable]{todonotes}
\usepackage{acronym}
\usepackage{tikz}
\usepackage{xspace}

\usetikzlibrary{positioning, arrows.meta}

\usepackage{amssymb}
\usepackage{xstring}
\usepackage{booktabs}

\newcommand \Description[2][] {}
\newcommand \posT {}

\newcommand{\citet}[1]{%
    \IfEqCase{#1}{%
            {kamalaruban2019interactive}{Kamalaruban et al. \cite{kamalaruban2019interactive}}%
            {yengera2021curriculum}{Yengera et al. \cite{yengera2021curriculum}}%
            {brown2018risk}{Brown et al. \cite{brown2018risk}}%
            {liu2017iterative}{Liu et al. \cite{liu2017iterative}}%
            {liu2018towards}{Liu et al. \cite{liu2018towards}}%
            {melo2018interactive}{Melo et al. \cite{melo2018interactive}}%
            {cakmak2012algorithmic}{Cakmak and Lopes \cite{cakmak2012algorithmic}}%
            {brown2019machine}{Brown and Niekum \cite{brown2019machine}}%
            {melo2021teaching}{Melo and Lopes \cite{melo2021teaching}}%
            {arora2021survey}{Arora and Doshi \cite{arora2021survey}}%
            {ziebart2013principle}{Ziebart et al. \cite{ziebart2013principle}}%
    }[\PackageError{tree}{Undefined option to tree: #1}{}]%
}%

\newcommand \tmul [2] {\langle #1, #2 \rangle}
\newcommand \phiVec {\boldsymbol \phi}
\newcommand \phiVecCe {\boldsymbol \phi^{\text{CE}}}
\newcommand \thVec   {\boldsymbol \theta}
\newcommand \rHat    {\hat{R}}
\newcommand \thVecHat{\hat{\boldsymbol \theta}}
\newcommand \piHat   {\hat{π}}
\newcommand \psiHat   {\hat{Ψ}}

\newcommand \piCe   {π^{\text{CE}}}
\newcommand \muVec  {\boldsymbol \mu}

\newcommand \expec  {\mathbb{E}}
\newcommand \proba  {\mathbb{P}}
\newcommand \reals  {\mathbb{R}}
\newcommand \trans  {\mathbb{T}}
\newcommand \Vso  {V^{\text{soft}}}
\newcommand \Qso  {Q^{\text{soft}}}
\newcommand \lambdaBold {\boldsymbol \lambda}
\newcommand \sQ {s^{\text{q}}}

\DeclareMathOperator {\firstS}{First-State}

\DeclareMathOperator {\svi}{Soft-Value-Iter}
\DeclareMathOperator {\imce}{Interactive-MCE}
\DeclareMathOperator {\dsr}{DSR}
\DeclareMathOperator {\ivar}{Interactive-VaR}
\DeclareMathOperator {\evd}{EVD}
\DeclareMathOperator {\softEvd}{Soft-EVD}
\DeclareMathOperator {\var}{VaR}
\DeclareMathOperator {\proj}{Proj}

\newcommand \algAgn {\textsc{Random}\xspace}
\newcommand \algRnd {\textsc{NoAL}\xspace}
\newcommand \algNoe {\textsc{Unmod}\xspace}
\newcommand \algVar {\textsc{TLimF}\xspace}

\newcommand \algCur {\textsc{TUnlimF}\xspace}

\DeclareUnicodeCharacter{00D7}{\times}
\DeclareUnicodeCharacter{00B7}{\cdot}
\DeclareUnicodeCharacter{1D62}{_{i}}
\DeclareUnicodeCharacter{2080}{_{0}}
\DeclareUnicodeCharacter{2081}{_{1}}
\DeclareUnicodeCharacter{2082}{_{2}}
\DeclareUnicodeCharacter{208A}{_{+}}
\DeclareUnicodeCharacter{2090}{_{a}}
\DeclareUnicodeCharacter{2095}{_{h}}
\DeclareUnicodeCharacter{2096}{_{k}}
\DeclareUnicodeCharacter{2099}{_{n}}
\DeclareUnicodeCharacter{209B}{_{s}}
\DeclareUnicodeCharacter{209C}{_{t}}
\DeclareUnicodeCharacter{2192}{\rightarrow}
\DeclareUnicodeCharacter{2190}{\leftarrow}
\DeclareUnicodeCharacter{2192}{\rightarrow}
\DeclareUnicodeCharacter{2200}{\forall}
\DeclareUnicodeCharacter{2203}{\exists}
\DeclareUnicodeCharacter{2207}{\nabla}
\DeclareUnicodeCharacter{2208}{\in}
\DeclareUnicodeCharacter{221E}{\infty}
\DeclareUnicodeCharacter{2260}{\neq}
\DeclareUnicodeCharacter{2264}{\leq}
\DeclareUnicodeCharacter{2265}{\geq}
\DeclareUnicodeCharacter{22C6}{^\star}
\DeclareUnicodeCharacter{2282}{\subset}
\DeclareUnicodeCharacter{2C7C}{_{j}}

\newcommand \acro[2]{\acrodef{#1}{#2}}

\acro{AL}              {Active Learning}
\acro{BEC}             {behavioral equivalence class}
\acro{CrossEnt-BC}     {Cross-Entropy Behavioral Cloning}
\acro{DBAL}            {Demonstration-based Active Learning}
\acro{DBAL-T}          {DBAL-based Teaching}
\acro{DSR}             {difficulty score ratio}
\acro{EVD}             {expected value difference}
\acro{HOV}             {high-occupancy vehicle}
\acro{IRL}             {Inverse Reinforcement Learning}
\acro{MCE}             {Maximum Causal Entropy}
\acro{MCE-IRL}         {Maximum Causal Entropy Inverse Reinforcement Learning}
\acro{MDP}             {Markov Decision Process}
\acro{MT}              {Machine Teaching}
\acro{NoAL}            {No Active Learning}
\acro{SCOT}            {Set Cover Optimal Teaching}
\acro{VaR}             {Value-at-Risk}
\acro{TLimF}           {Teaching with Limited Feedback}

\acused{MCE-IRL}
\acused{DBAL-T}

\newcommand \Act {\mathcal{A}}
\newcommand \Sta {\mathcal{S}}
\newcommand \Vpol {\mathcal{V}}

\ecaisubmission
\paperid{593}

\begin{document}

\begin{frontmatter}

\title{Interactively Teaching an Inverse Reinforcement Learner with Limited Feedback}

\author[A]{\fnms{Rustam}~\snm{Zayanov}\orcid{0009-0006-5301-6382}}
\author[B]{\fnms{Francisco}~\snm{S. Melo}\orcid{0000-0001-5705-7372}}
\author[B]{\fnms{Manuel}~\snm{Lopes}\orcid{0000-0002-6238-8974}}

\address[A]{Instituto Superior Técnico, Universidade de Lisboa}
\address[B]{INESC-ID \& Instituto Superior Técnico, Universidade de Lisboa}

\begin{abstract}

    We study the problem of teaching via demonstrations in sequential decision-making tasks.
    In particular, we focus on the situation when the teacher has no access to the learner's model and policy, and the feedback from the learner is limited to trajectories that start from states selected by the teacher.
    The necessity to select the starting states and infer the learner's policy creates an opportunity for using the methods of inverse reinforcement learning and active learning by the teacher.
    In this work, we formalize the teaching process with limited feedback and propose an algorithm that solves this teaching problem.
    The algorithm uses a modified version of the active value-at-risk method to select the starting states, a modified maximum causal entropy algorithm to infer the policy, and the difficulty score ratio method to choose the teaching demonstrations.
    We test the algorithm in a synthetic car driving environment and conclude that the proposed algorithm is an effective solution when the learner's feedback is limited.
\end{abstract}

\end{frontmatter}

\section{Introduction} \label{chap:intro}

\Acf{MT} is a computer science field that formally studies a learning process from a teacher's point of view.
The teacher's goal is to teach a target concept to a learner by demonstrating an optimal (often the shortest) sequence of examples.
\Ac{MT} has the potential to be applied to a wide range of practical problems \cite{zhu2015machine, zhu2018overview}, such as: developing better Intelligent Tutoring Systems for automated teaching for humans, developing smarter learning algorithms for robots, determining the teachability of various concept classes, testing the validity of human cognitive models, and cybersecurity.

One promising application domain of \ac{MT} is the automated teaching of sequential decision skills to human learners, such as piloting an airplane or performing a surgical operation.
In this domain, \ac{MT} can be combined with the theory of \acf{IRL} \cite{ng2000algorithms, abbeel2004apprenticeship}, also known as Inverse Optimal Control.
\Ac{IRL} formally studies algorithms for inferring an agent's goal based on its observed behavior in a sequential decision setting.
Assuming that a learner will use a specific \ac{IRL} algorithm to process the teacher's demonstrations, the teacher could pick an optimal demonstration sequence for that algorithm.

Most \ac{MT} algorithms assume that the teacher knows the learner's model, that is, the learner's algorithm of processing demonstrations and converting them into knowledge about the target concept.
In the case of human cognition, formalizing and verifying such learner models is still an open research question.
The scarcity of such models poses a challenge to the application of \ac{MT} to automated human teaching.
One way of alleviating the necessity of a fully defined learner model is to develop \ac{MT} algorithms that make fewer assumptions about the learner.
In the sequential decision-making domain, \citet{kamalaruban2019interactive} and \citet{yengera2021curriculum} have proposed teaching algorithms that admit some level of uncertainty about the learner model.
In particular, their teaching algorithms assume that the learner's behavior (policy) is maximizing some reward function, but it is unknown how the learner updates that reward function given the teacher's demonstrations.
To cope with this uncertainty, the teacher is allowed to observe the learner's behavior during the teaching process and infer the learner's policy from the observed trajectories, thus making the process iterative and interactive.

Both works assume that the teacher can periodically observe \textit{many} learner's trajectories from \textit{every initial state} and thus estimate the learner's policy with high precision.
Unfortunately, the need to produce many trajectories from every initial state may be unfeasible in real-life scenarios.
In our present work, we address a more realistic scenario in which the feedback from the learner is limited to just \textit{one trajectory} per each iteration of the teaching process.
The limit on the learner's feedback poses a challenge for the teacher in reliably estimating the learner's policy, which, in turn, may diminish the usefulness of the teacher's demonstrations.
Thus, our research question is:
\textbf{
    What are the effective ways of teaching an inverse reinforcement learner when the learner's policy and update algorithm are unknown, and the learner's feedback is limited?
}

The teacher's ability to precisely estimate the learner's policy greatly depends on the informativeness of the received trajectories.
We consider two scenarios: an unfavorable scenario when the teacher has no influence on what trajectories it will receive, and a more favorable scenario when the teacher can choose the states from which the learner will generate trajectories.
The necessity to select the starting states creates an opportunity for the teacher to use methods of \acf{AL} \cite{settles2009active}.
In the context of sequential decision-making, \Ac{AL} considers situations when a learner has to infer an expert's reward and can interactively choose the states from which the expert's demonstrations should start \cite{lopes2009active}.

The contribution of our work is two-fold.
Firstly, we propose a new framework that formalizes interactive teaching when the learner's feedback is limited.
Secondly, we propose an algorithm for teaching with limited feedback.
The algorithm performs three steps per every teaching iteration: selection of a query state (\ac{AL} problem), inference of the current learner's policy (\ac{IRL} problem), and selection of a teaching demonstration (\ac{MT} problem).
The algorithm uses a modified version of the Active-VaR \cite{brown2018risk} method for choosing query states, a modified version of the \ac{MCE} \cite{ziebart2013principle} method for inferring the learner's policy, and the \acf{DSR} \cite{yengera2021curriculum} method for selecting the teaching demonstration
We test the algorithm in a synthetic car driving environment and conclude that it is a viable solution when the learner's feedback is limited\footnotemark.

\footnotetext{The implementation of the algorithms is available at \url{https://github.com/rzayanov/irl-teaching-limited-feedback}}

\section{Related work} \label{chap:related}

\citet{liu2017iterative} explore the problem of \ac{MT} with unlimited feedback in the domain of supervised learning when the teacher and the learner represent the target concept as a linear model.
They consider a teacher that does not know the feature representation and the parameter of the learner.
For this scenario, they introduce an interaction protocol with unlimited learner feedback, where the teacher can query the learner at every step by sending all possible examples and receiving all learner's output labels.
\citet{liu2018towards} continue this work and explore teaching with limited feedback in the same supervised learning setting.
Similarly to our work, the teacher can not request all learner's labels at every step but instead has to choose which examples to query using an \ac{AL} method.

\citet{melo2018interactive} explore how interaction can help when the teacher has wrong assumptions about the learner.
The authors focus on the problem of teaching the learners that aim to estimate the mean of a Gaussian distribution given scalar examples.
When the teacher knows the correct learner model, the teaching goal is achieved after showing one example.
When it has wrong assumptions, and no interaction is allowed, the learner approaches the correct mean only asymptotically.
When interaction is allowed, the teacher can query the learner at any time, and the learner responds with the value of its current estimate perturbed by noise.
They show that this kind of interaction significantly boosts teaching progress.

\citet{cakmak2012algorithmic} and \citet{brown2019machine} propose non-interactive \ac{MT} algorithms for sequential decision-making tasks.
Both algorithms produce a minimal set of demonstrations that is sufficient to reliably infer the reward function.
Both algorithms are agnostic of the learner model and don't specify the order of demonstrations, which might be crucial for teaching performance if the learner is not capable of processing the whole set at once.
The algorithm of \citet{cakmak2012algorithmic} is based on the assumptions that the reward is a linear combination of \textit{state features} and that the teacher will provide enough demonstrations for the learner to estimate the teacher's expected \textit{feature counts} reliably.
With these assumptions, each demonstrated state-action pair induces a half-space constraint on the reward weight vector.
Assuming that the learner weights are bounded, it is possible to estimate the volume of the subspace defined by any set of such constraints.
A smaller volume means less uncertainty regarding the true weight vector.
Thus, demonstrations that minimize the subspace volume are preferred.
The authors propose a non-interactive algorithm for choosing the demonstration set: at every step, the teacher will pick a demonstration that minimizes the resulting subspace volume.
\citet{brown2019machine} propose an improved non-interactive \ac{MT} algorithm called \acf{SCOT}.
They first define a policy's \acfi{BEC} as a set of reward weights under which that policy is optimal.
A \ac{BEC} of a demonstration given a policy is the intersection of half-spaces formed by all state-action pairs present in such demonstration.
The authors propose finding the smallest set of demonstrations whose \ac{BEC} is equal to the \ac{BEC} of the optimal policy.
Finding such a set is a set-cover problem. 
The proposed algorithm is based on generating $m$ demonstrations from each starting state and using a greedy method of picking candidates.

\citet{kamalaruban2019interactive} and \citet{yengera2021curriculum} propose interactive \ac{MT} algorithms for sequential decision-making tasks when the learner can process only one demonstration at a time, but the feedback from the learner is unlimited or has a high limit.
\citet{kamalaruban2019interactive} first consider an omniscient teacher whose goal is to steer the learner toward the optimal weight parameter and find an effective teaching algorithm.
Next, they consider a less informative teacher that can not observe the learner's policy and has no information about the learner's feature representations and the update algorithm.
Instead of directly observing the current learner's policy $π^L_i$, the teacher can periodically request the learner to generate $k$ trajectories from \textit{every initial state}, thus estimating $π^Lᵢ$.
The limitation of this approach is that the necessity to produce trajectories from every initial state may be hard to implement in practice when $k$ is high or the number of initial states is high.
\citet{yengera2021curriculum} further explore the problem of teaching with unlimited feedback and propose the \acf{DSR} algorithm.
They introduce the notion of a \textit{difficulty score} of a trajectory given a policy, which is proportional to its conditional likelihood given that policy, and
propose a teaching algorithm that selects a trajectory that maximizes the ratio of difficulty scores of the learner's policy and the target policy.

To the best of our knowledge, the problem of teaching with limited feedback in the domain of sequential decision-making tasks has not yet been addressed in the literature.

\section{Problem formalism} \label{chap:formalism}

The underlying task to be solved by an agent is formally represented as a \acf{MDP} denoted as $M = (\Sta,\Act,\trans,\proba₀,γ,R⋆)$, where $\Sta$ is the set of states, $\Act$ is the set of actions, $\trans(S' \mid s,a)$ is the state transition probability upon taking action $a$ in state $s$, $\proba₀(S)$ is the initial state distribution, $γ$ is the discount factor, and $R⋆: \Sta → \reals$ is the reward function to be learned.

A stationary \textit{policy} is a mapping $π$ that maps each state $s ∈ \Sta$ into a probability distribution $π(\cdot \mid s)$ over $\Act$.
A policy can be executed in $M$, which will produce a sequence of state-action pairs called \textit{trajectory}.
For any trajectory $ξ=\{s₀,a₀,\ldots,s_T,a_T\}$, we will denote its $i$-th state and action as $s^ξᵢ$ and $a^ξᵢ$, respectively.
Given a policy $π$, the \textit{state-value function} $V^π(s)$, the \textit{expected policy value} $\Vpol^π$, and the \textit{Q-value function} $Q^π(s,a)$ are defined as follows respectively:
\begin{align}
    V^π(s)   &= \expec\left[ \sum_{t=0}^∞ γ^t R(Sₜ)  \mid  π, \trans,S₀ = s \right] \\
    \Vpol^π      &= \expec_{S \sim \proba₀}[V^π(S)] \\
    Q^π(s,a) &= R(s) + γ \expec_{S \sim \trans(· \mid s,a)}[V^π(S)]
\end{align}
A policy $π⋆$ is considered \textit{optimal} if it has the highest state-values for every state.
For any \ac{MDP}, at least one optimal policy exists, which can be obtained via the policy iteration method \cite{sutton2018reinforcement}.

\section{Framework for teaching with limited feedback}

In this section, we present our contributions: the framework for teaching with limited feedback and an algorithm for solving the problem of teaching with limited feedback.

We consider two entities that can execute policies on $M$: a \textit{teacher} with complete access to $M$ and a \textit{learner} that can access all elements of $M$ except the reward function, which we denote as $M \setminus R⋆$.
The teacher and the learner can interact with each other iteratively, with every iteration consisting of five steps described in Algorithm~\ref{alg:interction}.
In the first step, the teacher chooses a \textit{query state} $\sQᵢ$ and asks the learner to generate a trajectory starting from $\sQᵢ$.
We assume that the query states can only be selected from the set of initial states, i.e.,  $\sQᵢ ∈ \Sta₀ = \{s: \proba₀(s) > 0\}$.
In the second step, the learner generates a \textit{trajectory} $ξᵢ^L$ by executing its policy starting from $\sQᵢ$ and sends it back to the teacher.
In the third step, the teacher uses the learner's trajectory to update its estimate of the learner's current reward $\rHat^Lᵢ$ and policy $\piHat^Lᵢ$.
In the fourth step, the teacher demonstrates the optimal behavior by generating a trajectory $ξᵢ^T$, which we call a \textit{demonstration}, and sending it to the learner.
In the last step, the learner learns from the demonstration to update its reward $R^Lᵢ$ and policy $π^Lᵢ$.
The teaching process is terminated when the \textit{teaching goal} is achieved, in the sense defined below.

\begin{algorithm} \posT
    \caption{Framework for teaching with limited feedback}
    \label{alg:interction}
    \begin{algorithmic}[1]
        \For{$i = 1, \dots, ∞$}
            \State Teacher sends a \textit{query state} $\sQᵢ$ and requests a \textit{trajectory} starting from it
            \State Learner generates and sends a \textit{trajectory} $ξᵢ^L$
            \State Teacher updates its estimate of the learner's reward $\rHat^Lᵢ$ and policy $\piHat^Lᵢ$
            \State Teacher generates and sends a \textit{demonstration} $ξᵢ^T$
            \State Learner updates its reward $R^Lᵢ$ and policy $π^Lᵢ$
            \State Stop if the teaching goal is achieved
        \EndFor
    \end{algorithmic}
\end{algorithm}

We consider the problem described above from the perspective of a teacher that has limited knowledge about the learner.
We consider the following set of assumptions:
\begin{itemize}
    \item \textbf{Access to state features}: Both the teacher and the learner can observe the same $d$ numerical \textit{features} associated with every state, formalized as a mapping $\phiVec: \Sta → \reals^d$.
        The (discounted) \textit{feature counts} are defined for a trajectory $ξ$ or for a policy $π$ and a state $s$ as follows:
        \begin{align}
            \muVec(ξ)   &= \sumₜ γ^t \phiVec(sₜ) \\
            \muVec(π,s) &= \expec \left[ \sumₜ γ^t \phiVec(Sₜ) \mid π, S₀ = s \right]
        \end{align}
    \item \textbf{Rationality}: At every iteration, the learner maintains some reward mapping $Rᵢ^L$ and derives a stationary policy $πᵢ^L$ that is appropriate for $Rᵢ^L$, which it uses to generate trajectories.
    The exact method of deriving $πᵢ^L$ from $Rᵢ^L$ is unknown to the teacher.
    \item \textbf{Reward as a function of features}:
    As it is common in the \ac{IRL} literature, the learner represents the reward as a linear function of state features: $Rᵢ^{L}(s) = \tmul{\thVecᵢ}{\phiVec(s)} $, where the vector $\thVecᵢ$ is called the \textit{feature weights}.
    Furthermore, we assume that the true reward $R⋆$ can be expressed as a function of these features, i.e., $∃\thVec⋆$ s.t. $∀s, R⋆(s) = \tmul{\thVec⋆}{\phiVec(s)}$.
    \item \textbf{Learning from demonstrations}: Upon receiving a demonstration $ξᵢ^T$, the learner uses it to update its parameter $\thVec_{i+1}$ and thus its reward $R_{i+1}^L$.
    The exact method of updating $\thVec_{i+1}$ from $ξ^Tᵢ$ is unknown to the teacher.
\end{itemize}

There are different ways of evaluating the teacher's performance.
In general, some notion of numerical \textit{loss} $Lᵢ$ is defined for every step (also called the teaching risk), and the teacher's goal is related to the progression of that loss.
Similarly to the previous works, we will use a common definition of the loss as the \acfi{EVD}: $Lᵢ = \Vpol^{π⋆} - \Vpol^{\piHat^Lᵢ}$ \cite{abbeel2004apprenticeship, ziebart2010modeling}, when evaluated against the real reward $R⋆$, and define the teaching goal as achieving a certain loss threshold $ε$ in the lowest number of iterations.

\begin{figure}
    \centering
    \includegraphics[width=\columnwidth]{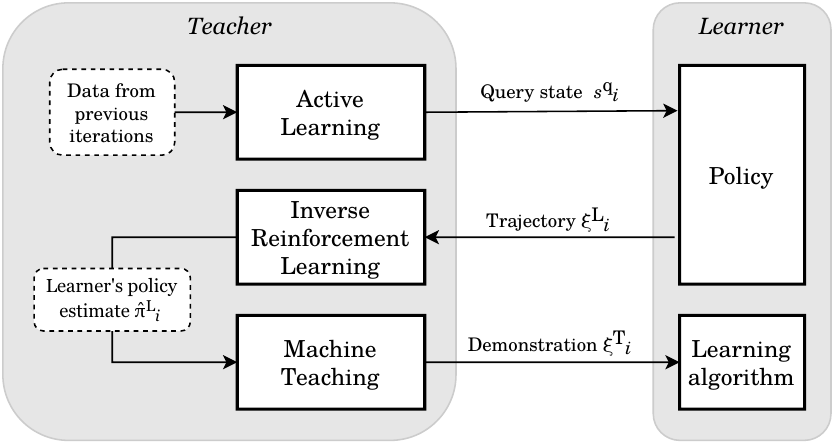}
    \caption{
        The teaching algorithm can be divided into three modules, each solving AL, IRL, or MT problem at every iteration.
        This diagram shows the inputs and outputs of the modules.
    }
    \Description[A flow diagram]{
        The diagram contains 3 rectangles, each representing a module of the teaching algorithm.
        The rectangles are connected with arrows to each other and to a circle representing the learner.
        The arrows have text explaining what data is passed between the modules and the learner.
    }
    \label{fig:diagram}
\end{figure}

Since the teacher has no access to $πᵢ^L$, it has to infer it from the trajectories received during teaching, which corresponds to the problem of \acl{IRL}.
To infer $πᵢ^L$ effectively, the teacher has to pick the query states with the highest potential of yielding an informative learner trajectory, which corresponds to the problem of \acl{AL}.
Finally, to achieve the ultimate goal of improving the learner's policy value, the teacher must select the most informative demonstrations to send, which corresponds to the problem of \acl{MT}.
Since a teaching algorithm has to solve these three problems sequentially, it can be divided into three ``modules'', each module solving one problem.
Figure~\ref{fig:diagram} shows the inputs and outputs of these modules.

We propose a concrete implementation of such a teaching algorithm, which we call \acf{TLimF}.
It is formally described in Algorithm~\ref{alg:tlimf}.
In the AL module, it uses the Interactive-\ac{VaR} algorithm, which is a version of the Active-\ac{VaR} algorithm \cite{brown2019machine} that we adapted to teaching with limited feedback.
In the IRL module, \ac{TLimF} uses the Interactive-\ac{MCE} algorithm, which is our adapted version of the \ac{MCE-IRL} algorithm \cite{ziebart2013principle}.
Finally, in the MT module, it uses the \ac{DSR} algorithm \cite{yengera2021curriculum}.
We describe all these three algorithms below.

\begin{algorithm} \posT
\caption{Teaching with Limited Feedback (TLimF)}
\label{alg:tlimf}
\begin{algorithmic}[1]
    \For{ $i = 1, \dots, ∞$ }
        \\\hrulefill

        \State $\sQᵢ = \ivar(ξ^L₁,\dots,ξ^Lᵢ, \piHat^L_{i-1})$
        \Comment \ac{AL} step
        \State Send $\sQᵢ$ to the learner, receive $ξ^Lᵢ$
        \\\hrulefill

        \State $\piHat^Lᵢ = \imce(ξ^Lᵢ, \thVecHat^L_{i-1})$
        \Comment \ac{IRL} step
        \\\hrulefill

        \State $ξ^Tᵢ = \dsr(\piHat^Lᵢ)$
        \Comment \ac{MT} step
        \State Send $ξ^Tᵢ$ to the learner
        \\\hrulefill

        \State Stop if $\Vpol^{π⋆} - \Vpol^{π^Lᵢ} < ε$
    \EndFor
\end{algorithmic}
\end{algorithm}

\subsection{Interactive-MCE}
Our algorithm for the \ac{IRL} module, Interactive-\ac{MCE}, is based on the \ac{MCE-IRL} algorithm proposed by \citet{ziebart2013principle}.
The original algorithm searches for a solution in the class \ac{MCE} policies,
\begin{align}
    π^{\thVec}(a \mid s) &= \exp [ β\Qso(s,a) - β\Vso(s) ], \\
    \Qso(s,a) &= \tmul{\thVec}{\phiVec(s)} + γ \expec_{S \sim \trans(· \mid s,a)}[\Vso(S)], \\
    \Vso(s)   &= \frac{1}{β} \log \sum_{a' ∈ \Act} \exp [ β\Qso(s,a')  ],
\end{align}
where $β$ is the entropy factor.
For any $\thVec$, the corresponding \ac{MCE} policy can be found with the soft-value iteration method \cite{ziebart2010modeling}.
The \ac{MCE-IRL} algorithm looks for a parameter $\thVec$ and a policy $π^{\thVec}$ that has the highest likelihood of producing the observed set of trajectories $Ξ$,
which is a convex problem when the reward is linear.
It can be solved with the gradient ascent method, with the gradient equal to
\begin{align}
    ∇L(\thVec) = \frac 1 {|Ξ|} \sum_ξ ( \muVec(ξ) - \muVec(π^{\thVec},s^ξ₀) ).
\end{align}

The original \ac{MCE-IRL} algorithm assumes that all the available trajectories were generated by a constant policy that is based on a constant reward function.
However, in our situation, the trajectories received from the learner are generated by different policies based on different rewards since the learner is assumed to update its reward function after receiving every teacher demonstration.
Thus, using all trajectories simultaneously with the \ac{MCE-IRL} algorithm might infer a reward that is very different from the actual learner's reward.

We propose a sequential version of this algorithm.
At every interaction step $i$, this algorithm starts with the previously inferred weights $\thVecHat_{i-1}$ and applies the \ac{MCE} gradient ascent with only the new trajectory $ξᵢ$ as the evidence.
Unlike a similar algorithm used by the \ac{MCE} learner in \cite{kamalaruban2019interactive}, which performs only one \ac{MCE} iteration per each new trajectory, our variant performs the gradient ascent for many iterations to better utilize the knowledge contained in the trajectories.
If the learner's trajectories are short, this method might overfit to the actions observed in the latest trajectory.
To avoid that, the older trajectories could be included in the gradient update, possibly with lower weight, or the feedback might have to be increased to a higher number of trajectories per iteration.
Interactive-MCE is formally described in Algorithm~\ref{alg:imce}.

\begin{algorithm} \posT
\caption{Interactive-MCE}
\label{alg:imce}
\begin{algorithmic}[1]
    \Require trajectory $ξᵢ$, previous or initial estimate $\thVecHat_{i-1}$
    \State $s₀ = \firstS(ξᵢ)$
    \State $\thVecHatᵢ = \thVecHat_{i-1}$
    \State $\piHatᵢ = \svi(\thVecHatᵢ)$
    \For{$n  = 1, \dots ,N$}
        \State $\thVecHatᵢ = \thVecHatᵢ + ηₙ(\muVec^{ξᵢ} - \muVec^{\piHatᵢ,s₀})$
        \State $\piHatᵢ = \svi(\thVecHatᵢ)$
    \EndFor
    \State \Return $\thVecHatᵢ, \piHatᵢ$
\end{algorithmic}
\end{algorithm}

\subsection{Interactive-VaR}
Our algorithm for the \ac{AL} module, Interactive-\ac{VaR}, is based on the Active-\ac{VaR} algorithm proposed by \citet{brown2019machine}.
The original algorithm assumes that the \ac{MDP} is deterministic, the reward weights lie on an L1-norm unit sphere, and the expert is following a constant parametrized \textit{softmax} policy,
\begin{align}
    π^{\thVec}(a \mid s) = \frac {\exp [ c Q^{\thVec}(s,a) ]} {\sum_{a' ∈ \Act} \exp [c Q^{\thVec}(s,a')]  },
\end{align}
where $c$ is a known confidence factor and $Q^{\thVec}$ are the Q-values of an optimal policy for $\thVec$.
For any reward weights $\thVec$ on the L1-norm unit sphere, the probability of observing the given set of trajectories $Ξ$ is
\begin{align}
    \proba(Ξ \mid \thVec) = \frac{1}{Z} \exp \left[\sum_{ξ∈Ξ} \sumₜ c Q^{\thVec}(s^ξₜ,a^ξₜ)\right],
\end{align}
where $Z$ is a normalizing constant.
If the apriori distribution of $\thVec$ is unknown, the probability of the given weights $\thVec$ generating the observed trajectories is
\begin{align}
    \proba(\thVec \mid Ξ) = \frac{1}{Z'} \proba(Ξ \mid \thVec).
\end{align}
For any policy $π$, weights $\thVec$ and starting state $s$, the \acf{EVD} of $π$ is defined as
\begin{align}
    \evd(\thVec \mid π,s) = \Vpol^π(s) - \Vpol^{π^{\thVec}}(s).
\end{align}
The Active-VaR method proposes to choose the next query state $\sQᵢ$ by finding the state that has the maximum \ac{VaR} of \ac{EVD} of the previously inferred policy $\piHat_{i-1}$:
\begin{align}
    \sQᵢ = \arg \max_{s ∈ \Sta₀} \var [\evd(\thVec \mid \piHat_{i-1},s)]
\end{align}

The original Active-VaR algorithm is not well-suited for the problem in question because the observations were generated by different learner policies, each corresponding to a different reward.
One way of addressing this problem is to give less weight to the older observations when computing the likelihood of any $\thVec$:
\begin{align}
    \proba(\thVec \mid ξ^L₁,\dots,ξ^Lₖ) & \approx \frac{1}{Z} \prod_{i=1}^k \proba(ξ^Lᵢ \mid \thVec)^{\lambdaBoldᵢ} \\
    \lambdaBoldₖ & = 1 \\
    \lim_{i → -∞} \lambdaBoldᵢ & = 0
\end{align}
In particular, it is possible to consider only the last $n$ observations,
\begin{align}
    \proba(\thVec \mid ξ^L₁,\dots,ξ^Lₖ) & \approx \frac{1}{Z} \prod_{i=(k-n)^+}^k \proba(ξ^Lᵢ \mid \thVec)
\end{align}
or to have the weight decay exponentially,
\begin{align}
    \proba(\thVec \mid ξ^L₁,\dots,ξ^Lₖ) & \approx \frac{1}{Z} \prod_{i=1}^k \proba(ξ^Lᵢ \mid \thVec)^{λ^{k-i}}, λ < 1.
\end{align}
An additional advantage of the exponential decay is computational speed because after receiving a new trajectory, it is possible to compute the updated likelihoods by reusing the likelihoods computed in the previous iteration:
\begin{align}
    \prod_{i=1}^k \proba(ξ^Lᵢ \mid \thVec)^{λ^{k-i}} = Pₖ = P_{k-1}^λ \proba(ξ^Lᵢ \mid \thVec).
\end{align}

To avoid using several policy classes within the compound teaching algorithm, we assume that the learner follows an \ac{MCE} policy instead of a softmax policy.
Given that, firstly, we use the soft Q-values of the \ac{MCE} policy to calculate the demonstration probabilities:
\begin{align}
    \proba(ξ|\thVec) = \frac{1}{Z} \exp \left[\sumₜ \Qso(sₜ,aₜ)\right]
\end{align}
Secondly, for calculating \ac{VaR}, we use the difference of the expected soft values: $\softEvd(\thVec|π,s) = V^{\text{soft}, π}(s) - V^{\text{soft}, π^{\thVec}}(s)$.
Finally, we replace the assumption about the known softmax confidence factor $c$ with a similar assumption about the known \ac{MCE} entropy factor $β$.

Interactive-MCE is formally described in Algorithm~\ref{alg:ivar}.
\begin{algorithm} \posT
\caption{Interactive Value-at-Risk (Interactive-VaR)}
\label{alg:ivar}
\begin{algorithmic}[1]
    \Require previous trajectories $ξ₁,\dots,ξᵢ$, previous or initial estimate $\piHat^L_{i-1}$
    \If {$i = 1$}
        \State Pick a random initial state $\sQᵢ ∈ \Sta₀$
    \Else
        \State Sample reward weights $Θ$
        \State $\sQᵢ = \arg \max_{s ∈ \Sta₀} \var [\softEvd(Θ \mid \piHat^L_{i-1}, s)]$
    \EndIf
    \State \Return $\sQᵢ$
\end{algorithmic}
\end{algorithm}

\subsection{Difficulty Score Ratio}
For deterministic MDPs, the \textit{difficulty score} of a demonstration $ξ$ w.r.t.\ a policy $π$ is defined as
\begin{align}
    Ψ(ξ) = \frac{1}{\prodₜ π(aₜ \mid sₜ)}
\end{align}
The \ac{DSR} algorithm selects the next teacher's demonstration $ξ^Tᵢ$ by iterating over a pool of candidate trajectories $Ξ$ and finding the trajectory with the maximum \acl{DSR}.
\ac{DSR} is formally described in Algorithm~\ref{alg:dsr}.

\begin{algorithm} \posT
\caption{Difficulty Score Ratio (DSR)}
\label{alg:dsr}
\begin{algorithmic}[1]
    \Require Policy estimate $\piHat^Lᵢ$
    \For{ $ξ$ in candidate pool $Ξ$ }
        \State $\psiHat^Lᵢ(ξ) = \prodₜ \piHat^Lᵢ(a^ξₜ \mid s^ξₜ)^{-1}$
        \State $Ψ^T(ξ) = \prodₜ π⋆(a^ξₜ \mid s^ξₜ)^{-1}$
    \EndFor
    \State $ξ^Tᵢ = \arg \max_{ξ∈Ξ} \frac{\psiHat^Lᵢ(ξ)}{Ψ^T(ξ)}$
    \State \Return $ξ^Tᵢ$
\end{algorithmic}
\end{algorithm}

\section{Experimental evaluation} \label{chap:experiments}

\begin{figure}
    \centering
    \includegraphics[width=.97\columnwidth]{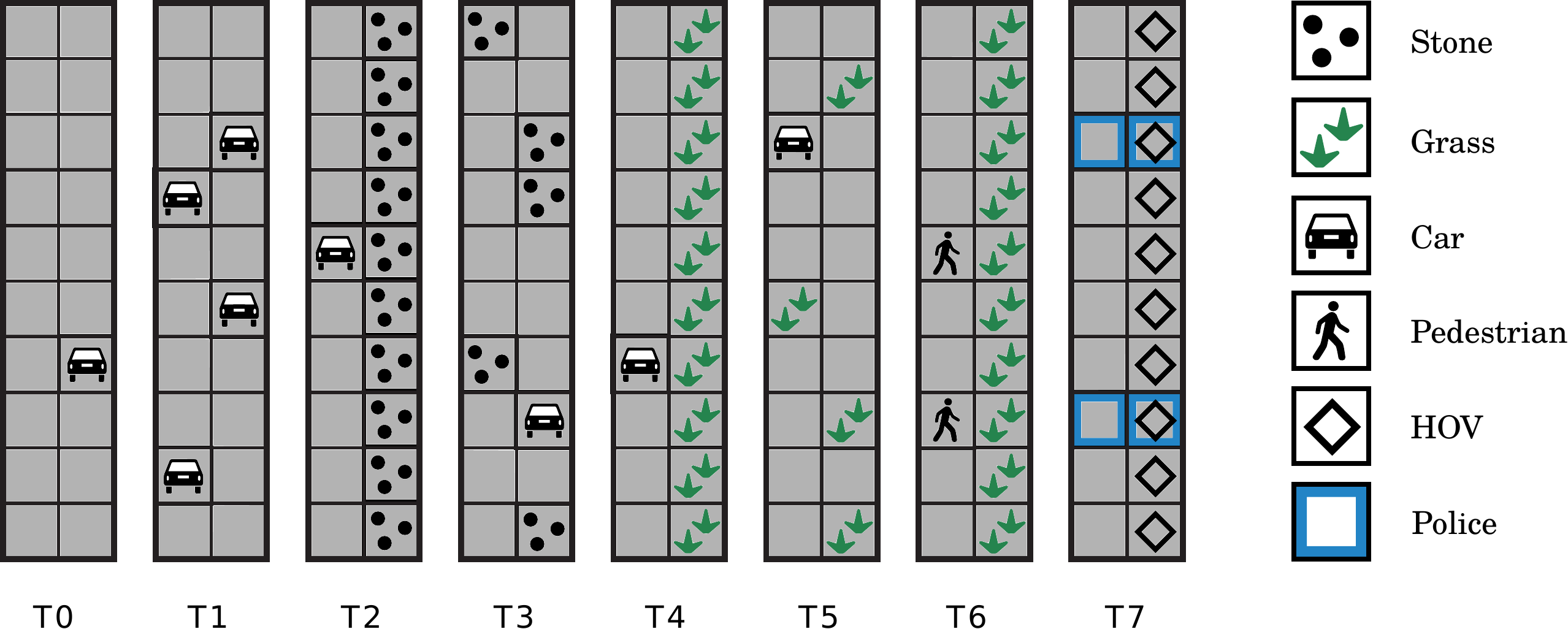}
    \caption{
        Examples of each road type of the car environment.
        Each $2 \times 10$ grid represents a road.
        The agent starts at the bottom left corner of a randomly selected road.
        After the agent has advanced for 10 steps upwards along the road, the MDP is terminated.
    }
    \Description[Eight grids of size 2 by 10]{Some grid cells contain features, such as cars, stones, grass, pedestrians, HOV and police. }
    \label{fig:roads}
\end{figure}

We tested our teaching algorithm in the synthetic car driving environment proposed by \citet{kamalaruban2019interactive}.
The environment consists of 40 isolated roads, each road having two lanes.
The agent represents a car that is driving along one of the roads.
The road is selected randomly at the start of the decision process, and the process terminates when the agent has reached the end of the road.
There are eight road types, with five roads of each type.
The road types, which we refer to as T0-T7, represent various driving conditions:
\begin{itemize}
    \item T0 roads are mostly empty and have a few other cars.
    \item T1 roads are more congested and have many other cars.
    \item T2 roads have stones on the right lane, which should be avoided.
    \item T3 roads have cars and stones placed randomly.
    \item T4 roads have grass on the right lane, which should be avoided.
    \item T5 roads have cars and grass placed randomly.
    \item T6 roads have grass on the right lane and pedestrians placed randomly, both of which should be avoided.
    \item
        T7 roads have a \ac{HOV} lane on the right and police at certain locations.
        Driving on a \ac{HOV} lane is preferred, whereas the police is neutral.
\end{itemize}
Each road is represented as a $2 \times 10$ grid.
We assume without loss of generality that only the agent is moving, other objects being static.
Roads of the same type differ in the placement of the random objects.
Figure~\ref{fig:roads} demonstrates example roads of all types.

The agent has three actions at every state: \texttt{left}, \texttt{right}, and \texttt{stay}.
Choosing \texttt{left} moves the agent to the left lane if it was on the right lane, otherwise moves it to a random lane.
Choosing \texttt{right} yields a symmetrical transition.
Choosing \texttt{stay} keeps the agent on the same lane.
Regardless of the chosen action, the agent always advances along the road.
The environment has 40 possible initial states, each corresponding to the bottom left corner of every road.
After advancing along the road for ten steps, the \ac{MDP} is terminated.
We assume $γ = 0.99$.

For every state, eight binary features are observable.
Six of them represent the environment objects: stone, grass, car, pedestrian, HOV, and police.
The last two indicate whether there's a car in the next cell or a pedestrian.
We consider the reward to be a linear function of these binary features, with reward weights specified in Table~\ref{tab:weights}.

\begin{table} \posT
    \centering
    \caption{True feature weights}
    \label{tab:weights}
    \begin{tabular}{rll}
        \toprule
        \textit{Feature} & \textit{Weight} \\
        \midrule
            \texttt{stone} & -1 \\
            \texttt{grass} & -0.5 \\
            \texttt{car} & -5 \\
            \texttt{pedestrian} & -10 \\
            \texttt{HOV} & +1 \\
            \texttt{police} & 0 \\
            \texttt{car-in-front} & -2 \\
            \texttt{ped-in-front} & -5 \\
        \bottomrule
    \end{tabular}
\end{table}

\subsection{CrossEnt-BC learner}
We use a linear variant of the \acf{CrossEnt-BC} learner proposed by \citet{yengera2021curriculum} for our experiments.
This learner follows a parametrized softmax policy,
\begin{align}
    \piCeᵢ(a \mid s) = \frac {\exp [ Hᵢ(s,a) ]} {\sum_{a' ∈ \Act} \exp [Hᵢ(s,a')]  },
\end{align}
where $Hᵢ$ is a parametric scoring function that depends on a parameter $\thVec^Lᵢ$ and a constant feature mapping,
\begin{align}
    \phiVecCe(s,a) = \expec_{S' \sim \trans(· \mid s,a)} [ \phiVec(S') ],
\end{align}
and is defined as $Hᵢ(s,a) = \tmul{\thVec^Lᵢ}{\phiVecCe(s,a)}$.
The likelihood of any demonstration $ξ$ and its gradient are defined respectively as
\begin{align}
    L(\thVec^Lᵢ) &= \log \proba(ξ \mid \thVec^Lᵢ), \\
    ∇L(\thVec^Lᵢ) &= \sumₜ \left( \phiVecCe(s^ξₜ, a^ξₜ) - \expec_{a \sim \piCeᵢ(·|s^ξₜ)}\left[\phiVecCe(s^ξₜ, a)\right] \right).
\end{align}

This learner starts with random initial weights $\thVec^L₁$, every element being uniformly sampled from $(-10, 10)$.
Upon receiving a new demonstration $ξ^Tᵢ$ from the teacher, the learner performs a projected gradient ascent,
\begin{align}
    \thVec^L_{i+1} = \proj_Θ \left[\thVec^Lᵢ + η ∇L(\thVec^Lᵢ)\right],
\end{align}
where $η = 0.34$ and $Θ$ is a hyperball centered at zero with a radius of 100.

\subsection{Teaching algorithms}

We compared the following algorithms, also presented in Table~\ref{tab:algos}:
\begin{itemize}
    \item \algAgn teacher does not infer $\thVec^Lᵢ$ and selects demonstrations by choosing a random initial state and generating an optimal demonstration from that state.
          This algorithm was originally proposed in \cite{kamalaruban2019interactive} and serves as the worst-case baseline.
    \item \algRnd teacher selects query states randomly but uses \ac{MCE} to infer the learner reward and DSR to select demonstrations.
        We included this algorithm as the second worst-case baseline to verify whether the usage of an \ac{AL} algorithm by other teachers can boost the teaching process.
    \item \algNoe uses unmodified Active-VaR to select query states, followed by unmodified MCE-IRL and DSR.
        We included this algorithm to verify whether the changes that we introduced in Interactive-VaR and Interactive-MCE affect the performance.
    \item \algVar uses Interactive-VaR to select query states, followed by Interactive-MCE and DSR.
    \item \algCur teacher knows the exact learner's policy at every step and therefore does not need \ac{AL} and \ac{IRL} modules.
        It uses the DSR algorithm to select demonstrations.
        This algorithm was originally proposed in \cite{yengera2021curriculum} and serves as the best-case baseline.
\end{itemize}

We did not include non-interactive algorithms in the experiment, because it was shown in \cite{yengera2021curriculum} that the state-of-the-art non-interactive \ac{MT} algorithm, \acl{SCOT}, did not perform better than \algAgn in this environment.
We also did not include the Black-Box (BBox) algorithm of \citet{kamalaruban2019interactive} in the comparison, because it was shown in \cite{yengera2021curriculum} that \algCur has similar performance to BBox and can be considered an improvement over it.

\begin{table} \posT
    \centering
    \caption{Tested algorithms}
    \label{tab:algos}
    \begin{tabular}{cccc}
        \toprule
        \textit{Name} & \textit{AL} & \textit{IRL} & \textit{MT} \\
        \midrule
            \algAgn & - & - & Random \\
            \algRnd & Random & Interactive-MCE & DSR \\
            \algNoe & Active-VaR & MCE-IRL & DSR \\
            \algVar & Interactive-VaR & Interactive-MCE & DSR \\
            \algCur & Not needed & Not needed & DSR \\
        \bottomrule
    \end{tabular}
\end{table}

The Interactive-VaR algorithm samples reward weights from the L1-norm sphere with a radius equal to 24, which is the L1-norm of the true feature weights.
The \ac{VaR} is computed on 5,000 uniformly sampled weights on the sphere\footnotemark.
\footnotetext{Increasing the sample size or sampling with MCMC yields similar results.}
For computing the posterior likelihood of $\thVec$, the demonstrations are weighted exponentially with $λ = 0.4$.
The \ac{EVD} between the two policies is computed using soft policy values.
The $α$ factor of \ac{VaR} was set to 0.95.
The MCE algorithms use 100 iterations of the gradient ascent.
The DSR algorithm selects demonstrations from a constant pool that consists of 10 randomly sampled trajectories per road.

\subsection{Analysis of the teacher's performance}

We conducted the experiment 16 times with different random seeds, which affected the random placement of objects on the roads and the random initial weights of the learners, and averaged the results of 16 experiments.

\begin{figure}
    \centering
    \begin{subfigure}[b]{.47\columnwidth}
        \centering
        \includegraphics[width=.97\columnwidth]{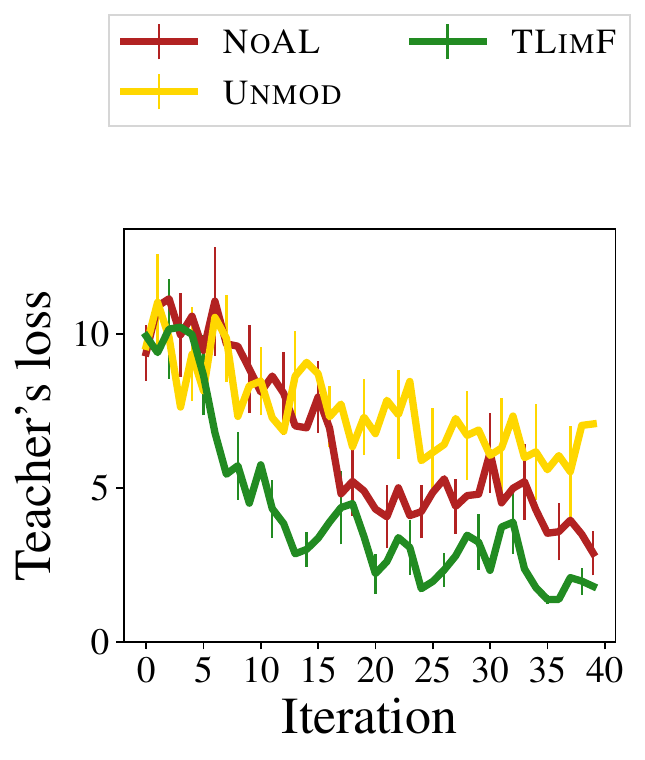}
        \caption{Teacher's inferred policy loss.}
        \Description[A line plot with 3 lines representing the teacher's losses for NoAL, Unmod, and TLimF.]{
            The X-axis measures the teaching iteration, the Y-axis measures the loss.
        }
        \label{fig:teacher_losses}
    \end{subfigure}
    \hfill
    \begin{subfigure}[b]{.47\columnwidth}
        \centering
        \includegraphics[width=.97\columnwidth]{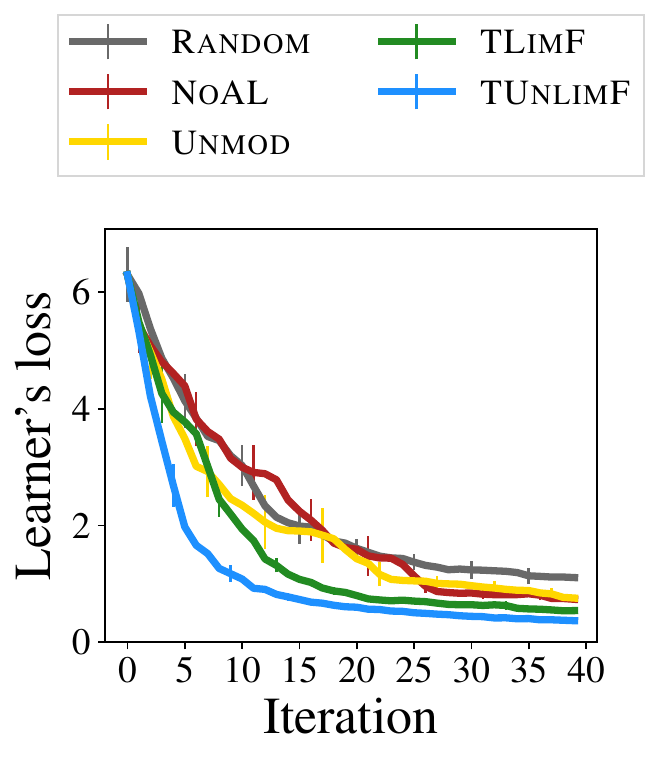}
        \caption{Learner's policy loss.}
        \Description[A line plot with 5 lines representing the learner's losses for Random, NoAL, Unmod, TLimF, and TUnlimF.]{
            The X-axis measures the teaching iteration, the Y-axis measures the loss.
        }
        \label{fig:losses}
    \end{subfigure}
    \captionsetup{subrefformat=parens}
    \caption{
        Teaching results are measured as \subref{fig:teacher_losses} the loss of the teacher's inferred policy and \subref{fig:losses} the loss of the actual learner's policy.
        The thick lines represent the averages of 16 experiments.
        The thin vertical lines measure the standard error.
        \algVar demonstrates the lowest losses during most of the process, with \algRnd and \algNoe significantly lagging behind.
    }
    \label{fig:results}
\end{figure}

Figure~\ref{fig:teacher_losses} displays the ability of the teaching algorithms to accurately estimate the current learner's policy.
For every iteration step, it shows the loss of the teacher's inferred policy $\piHat^Lᵢ$ w.r.t.\ the actual learner's policy $π^Lᵢ$.
The thick lines represent the average of 16 experiments, and the thin vertical lines measure the standard error.
As we can see, at any iteration, \algVar is able to estimate the learner's policy more reliably than \algRnd, which means that using an \ac{AL} algorithm is crucial for effectively estimating the learner's policy.
We can also see that \algNoe performs considerably worse than \algRnd, which means that unmodified Active-\ac{VaR} and \ac{MCE-IRL} algorithms are not suitable for teaching with limited feedback.
The teacher's performance in estimating the learner's policy is an intermediate result that affects the overall teaching performance, which is discussed next.

Figure~\ref{fig:losses} and table~\ref{tab:times} display the effectiveness of the teacher's effort in teaching the learner.
For every iteration step, figure~\ref{fig:losses} shows the loss of the learner's policy $π^Lᵢ$ w.r.t.\ the optimal policy $π⋆$, averaged over 16 experiments.
Table~\ref{tab:times} shows how many iterations, on average, the teachers need before reaching various loss thresholds.
As we can see, \algVar does not attain the performance of the upper baseline, \algCur, but it performs considerably better than other teaching algorithms: its loss is consistently lower starting from the seventh iteration, and it needs considerably fewer iterations to reach the presented loss thresholds.
This implies that \textit{for teaching with limited feedback, the best performance is achieved when the teacher is using specialized \ac{AL} and \ac{IRL} algorithms to select query states and infer the learner's policy}.
The \algRnd teacher performs worse than \algVar but better than the lower baseline, \algAgn: its loss is considerably lower starting from the 25th iteration, and it needs fewer iterations to reach the loss thresholds.
This implies that \textit{teaching without \ac{AL} is still better than selecting demonstrations randomly}.
Finally, the performance of the teacher with unmodified \ac{AL} and \ac{IRL} algorithms, \algNoe, is high during the first six iterations, but it gradually worsens during the teaching process and falls below the performance of \algRnd.
It also shows significantly low performance at reaching the loss thresholds, needing more iterations than the random teacher, which implies that modifying the algorithms was necessary for good performance.

\begin{table} \posT
    \centering
    \caption{Iterations needed to reach a loss threshold $ε$}
    \label{tab:times}
    \begin{tabular}{cccc}
        \toprule
        \textit{Teacher} & $ε=2$ & $ε=1$ & $ε=0.5$ \\
        \midrule
        \algAgn & 16 & 36 & 96 \\
        \algRnd & 13 & 25 & 49 \\
        \algNoe & 31 & 54 &102 \\
        \algVar &  8 & 19 & 46 \\
        \algCur &  5 & 11 & 27 \\
        \bottomrule
    \end{tabular}
\end{table}

\section{Summary and future work} \label{chap:summary}

We have proposed a teacher-learner interaction framework in which the feedback from the learner is limited to just one trajectory per teaching iteration.
Such a framework is closer to real-life situations and more challenging when compared with the frameworks used in previous works.
In this framework, the teacher has to solve \ac{AL}, \ac{IRL}, and \ac{MT} problems sequentially at every teaching iteration.
We have proposed a teaching algorithm that consists of three modules, each dedicated to solving one of these three sub-problems.
This algorithm uses a modified \ac{MCE-IRL} algorithm for solving the \ac{IRL} sub-problem, a modified Active-\ac{VaR} algorithm for solving the \ac{AL} problem, and the \ac{DSR} algorithm for solving the \ac{MT} problem.
We have tested the algorithm on a synthetic car-driving environment and compared it with the existing algorithms and the worst-case baseline.
We have concluded that the new algorithm is effective at solving the teaching problem.

In future work, it would be interesting to study such a teacher-learner interaction in more complex environments.
For example, an environment could have more states and a non-linear reward function possibly represented as a neural network.
Another question yet to be addressed is the convergence guarantees of the proposed algorithms.
It is also interesting to check whether the \ac{MT} module of the algorithm could be improved by considering the uncertainty of the estimated learner policy.
Another possible direction of research is finding more sophisticated ways of weighing older trajectories of the learner.
E.g., if the environment consists of several isolated regions and any feature is confined to a certain region, then sending a teaching demonstration in one region might not change the learner's behavior in others, therefore the previous learner's trajectories from other regions might not need to be weighed down.

\balance


\begin{thebibliography}{10}

\bibitem{abbeel2004apprenticeship}
Pieter Abbeel and Andrew~Y Ng, `Apprenticeship learning via inverse
  reinforcement learning', in {\em Proceedings of the twenty-first
  international conference on Machine learning}, p.~1, (2004).

\bibitem{brown2018risk}
Daniel~S Brown, Yuchen Cui, and Scott Niekum, `Risk-aware active inverse
  reinforcement learning', in {\em Conference on Robot Learning}, pp. 362--372.
  PMLR, (2018).

\bibitem{brown2019machine}
Daniel~S Brown and Scott Niekum, `Machine teaching for inverse reinforcement
  learning: Algorithms and applications', in {\em Proceedings of the AAAI
  Conference on Artificial Intelligence}, volume~33, pp. 7749--7758, (2019).

\bibitem{cakmak2012algorithmic}
Maya Cakmak and Manuel Lopes, `Algorithmic and human teaching of sequential
  decision tasks', in {\em Twenty-Sixth AAAI Conference on Artificial
  Intelligence}, (2012).

\bibitem{kamalaruban2019interactive}
Parameswaran Kamalaruban, Rati Devidze, Volkan Cevher, and Adish Singla,
  `Interactive teaching algorithms for inverse reinforcement learning', {\em
  arXiv preprint arXiv:1905.11867}, (2019).

\bibitem{liu2017iterative}
Weiyang Liu, Bo~Dai, Ahmad Humayun, Charlene Tay, Chen Yu, Linda~B Smith,
  James~M Rehg, and Le~Song, `Iterative machine teaching', in {\em
  International Conference on Machine Learning}, pp. 2149--2158. PMLR, (2017).

\bibitem{liu2018towards}
Weiyang Liu, Bo~Dai, Xingguo Li, Zhen Liu, James Rehg, and Le~Song, `Towards
  black-box iterative machine teaching', in {\em International Conference on
  Machine Learning}, pp. 3141--3149. PMLR, (2018).

\bibitem{lopes2009active}
Manuel Lopes, Francisco Melo, and Luis Montesano, `Active learning for reward
  estimation in inverse reinforcement learning', in {\em Joint European
  Conference on Machine Learning and Knowledge Discovery in Databases}, pp.
  31--46. Springer, (2009).

\bibitem{melo2018interactive}
Francisco~S Melo, Carla Guerra, and Manuel Lopes, `Interactive optimal teaching
  with unknown learners.', in {\em IJCAI}, pp. 2567--2573, (2018).

\bibitem{ng2000algorithms}
Andrew~Y Ng, Stuart Russell, et~al., `Algorithms for inverse reinforcement
  learning.', in {\em Icml}, volume~1, p.~2, (2000).

\bibitem{settles2009active}
Burr Settles, `Active learning literature survey', Computer Sciences Technical
  Report 1648, University of Wisconsin--Madison, (2009).

\bibitem{sutton2018reinforcement}
Richard~S Sutton and Andrew~G Barto, {\em Reinforcement learning: An
  introduction}, MIT press, 2018.

\bibitem{welch1947generalization}
Bernard~L Welch, `The generalization of ‘student's’problem when several
  different population varlances are involved', {\em Biometrika}, {\bf
  34}(1-2),  28--35, (1947).

\bibitem{yengera2021curriculum}
Gaurav Yengera, Rati Devidze, Parameswaran Kamalaruban, and Adish Singla,
  `Curriculum design for teaching via demonstrations: Theory and applications',
  {\em Advances in Neural Information Processing Systems}, {\bf 34},
  10496--10509, (2021).

\bibitem{zhu2015machine}
Xiaojin Zhu, `Machine teaching: An inverse problem to machine learning and an
  approach toward optimal education', in {\em Proceedings of the AAAI
  Conference on Artificial Intelligence}, volume~29, (2015).

\bibitem{zhu2018overview}
Xiaojin Zhu, Adish Singla, Sandra Zilles, and Anna~N Rafferty, `An overview of
  machine teaching', {\em arXiv preprint arXiv:1801.05927}, (2018).

\bibitem{ziebart2010modeling}
Brian~D Ziebart, {\em Modeling purposeful adaptive behavior with the principle
  of maximum causal entropy}, Carnegie Mellon University, 2010.

\bibitem{ziebart2013principle}
Brian~D Ziebart, J~Andrew Bagnell, and Anind~K Dey, `The principle of maximum
  causal entropy for estimating interacting processes', {\em IEEE Transactions
  on Information Theory}, {\bf 59}(4),  1966--1980, (2013).

\end{thebibliography}
\end{document}